\documentclass[10pt,twocolumn,letterpaper]{article}

 \usepackage{cvpr}              % To produce the CAMERA-READY version

\usepackage{amsmath,bm}
\definecolor{cvprblue}{rgb}{0.21,0.49,0.74}
\usepackage[pagebackref,breaklinks,colorlinks,allcolors=cvprblue]{hyperref}
\usepackage{bbding}
\usepackage[flushleft]{threeparttable}
\def\paperID{31808} % *** Enter the Paper ID here
\def\confName{CVPR}
\def\confYear{2026}

\title{CADC: Content Adaptive Diffusion-Based Generative Image Compression}

\author{
Xihua Sheng$^{1}$ , Lingyu Zhu$^{1}$, Tianyu Zhang$^{2}$, Dong Liu$^{2}$, Shiqi Wang$^{1}$, Jing Wang$^{3}$\thanks{Corresponding author} \\
$^{1}$City University of Hong Kong  $^{2}$University of Science and Technology of China  \\ $^{3}$Central Media Technology  Institute, Huawei \\
{\tt\small \{xihsheng, shiqwang\}@cityu.edu.hk, lingyzhu-c@my.cityu.edu.hk,}\\
{\tt\small  zhangtianyu@mail.ustc.edu.cn, dongeliu@ustc.edu.cn, wangjing215@huawei.com}
}

\begin{document}
\twocolumn[{%
\renewcommand\twocolumn[1][]{#1}%
\maketitle

\begin{center}
    \centering
    \captionsetup{type=figure}
    \includegraphics[width=\textwidth]{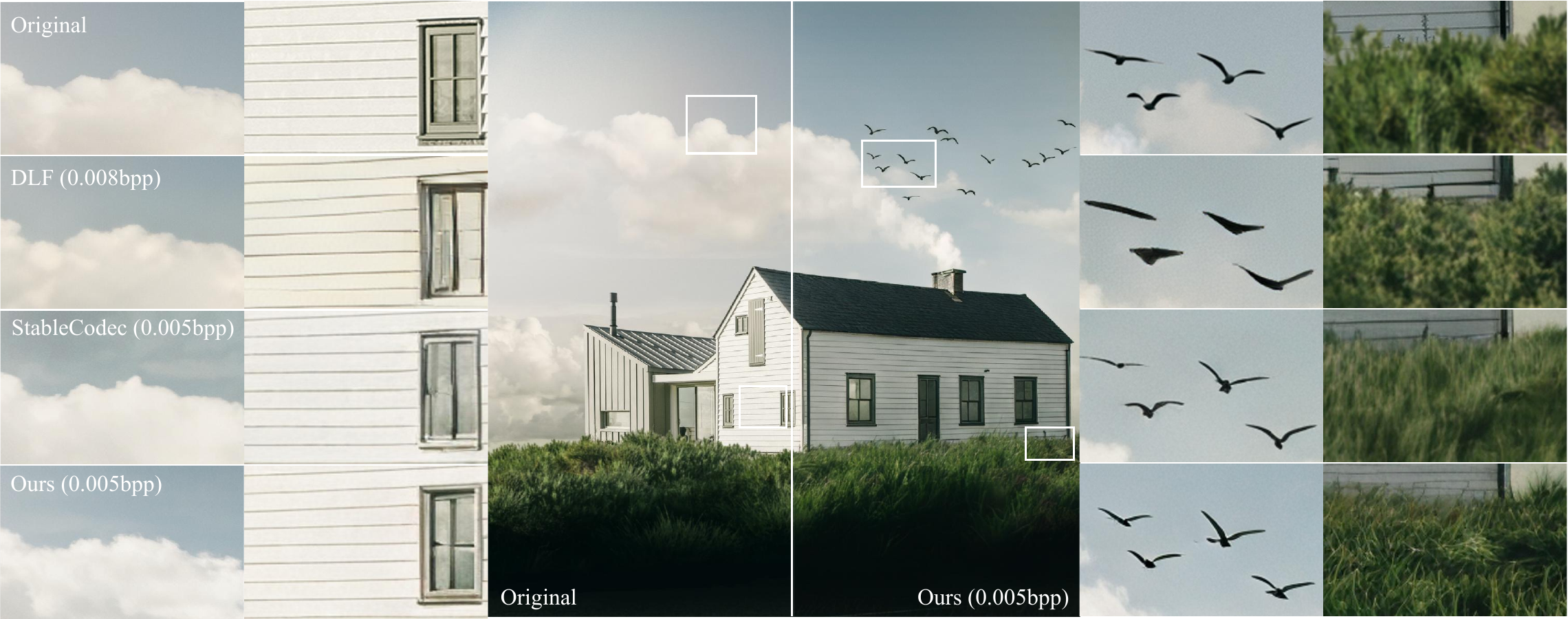}
    \captionof{figure}{A qualitative comparison between our codec, StableCodec~\cite{zhang2025stablecodec}, and DLF~\cite{xue2025dlf} when compressing a 2K-resolution image of the test set of CLIC 2020 Professional~\cite{toderici2020clic} under ultra-low bitrate conditions. Our codec produces images with high visual quality, especially in regions with complex texture. In contrast, DLF and StableCodec exhibit noticeable artifacts, such as blurring and color shifting.}
  \label{fig:subjective2}
\end{center}%
}]
\let\thefootnote\relax\footnotetext{* Corresponding author}
\begin{abstract}
Diffusion-based generative image compression has demonstrated remarkable potential for achieving realistic reconstruction at ultra-low bitrates. The key to unlocking this potential lies in making the entire compression process content-adaptive, ensuring that the encoder's representation and the decoder's generative prior are dynamically aligned with the semantic and structural characteristics of the input image. However, existing methods suffer from three critical limitations that prevent effective content adaptation. First, isotropic quantization applies a uniform quantization step, failing to adapt to the spatially varying complexity of image content and creating a misalignment with the diffusion model's noise-dependent prior.~Second, the information concentration bottleneck---arising from the dimensional mismatch between the high-dimensional noisy latent and the diffusion decoder's fixed input---prevents the model from adaptively preserving essential semantic information in the primary channels.~Third, existing textual conditioning strategies either need significant textual bitrate overhead or rely on generic, content-agnostic textual prompts, thereby failing to provide adaptive semantic guidance efficiently.
To overcome these limitations, we propose a content-adaptive diffusion-based image codec (CADC) with three technical innovations: 1) an Uncertainty-Guided Adaptive Quantization (UGAQ) method that learns spatial uncertainty maps to adaptively align quantization distortion with content characteristics; 2) an Auxiliary Decoder-Guided Information Concentration (ADGIC) method that uses a lightweight auxiliary decoder to enforce content-aware information preservation in the primary latent channels; and 3) a Bitrate-Free Adaptive Textual Conditioning (BFATC) method that derives content-aware textual descriptions from the auxiliary reconstructed image, enabling semantic guidance without bitrate cost. Comprehensive experimental results show that our codec achieves state-of-the-art perceptual quality at ultra-low bitrates.

\end{abstract}    

\section{Introduction}
\label{sec:intro}
Digital images account for a substantial portion of internet traffic, driving continuous demand for efficient compression technologies that reduce storage and transmission costs while maintaining visual quality. Traditional image codecs~\cite{wallace1991jpeg,skodras2002jpeg,sullivan2012overview,bross2021overview} have been widely adopted for decades. To pursue higher compression performance, learned image compression~\cite{DBLP:conf/iclr/BalleLS17,cheng2020image,feng2025linear,zou2022devil,lu2021transformer,liu2023learned,agustsson2017soft,guo2021soft,agustsson2020universally,ge2024nlic,DBLP:conf/iclr/BalleMSHJ18,DBLP:conf/nips/MinnenBT18,minnen2020channel,he2021checkerboard,he2022elic,fu2024fast,jiang2023mlic,qian2022entroformer,li2024groupedmixer} has emerged in recent years. However, at ultra-low bitrates, even state-of-the-art learned image codecs tend to produce reconstructions plagued by blurring and loss of fine details, as they primarily optimize for pixel-level signal fidelity rather than perceptual quality~\cite{blau2019rethinking}.\par
Generative image compression addresses this limitation by leveraging strong generative models to produce visually realistic reconstructions.~Existing generative image codecs can be broadly categorized into three classes: Generative Adversarial Network (GAN)-based codecs~\cite{mentzer2020high,agustsson2023multi,li2022content,agustsson2019generative,chen2023compact}, which employ adversarial training~\cite{goodfellow2020generative} to enhance visual quality; vector quantization (VQ)-based codecs~\cite{xue2025one,jia2024generative,qi2025generative}, which learn discrete representations for efficient compression; and diffusion-based codecs~\cite{careil2023towards,gao2025exploring,zhang2025stablecodec,xue2025one,relic2025bridging,relic2024lossy,li2024diffusion,li2024towards,ke2025ultra,kuang2024consistency}, which utilize diffusion denoising to reconstruct images. While GAN-based and vector quantization codecs have shown promising results, their generative capabilities remain inherently constrained. In contrast, diffusion-based codecs---benefiting from the exceptional generative capacity of diffusion models---have recently achieved remarkable performance, particularly under ultra-low bitrate conditions. \par
Despite these advances, we identify three fundamental limitations in current diffusion-based image codecs that prevent them from achieving effective content adaptation, which is crucial for optimally leveraging generative priors across diverse image structures and semantics. First, the prevalent use of \textit{isotropic quantization} applies a uniform quantization step across the compact latent representation, ignoring the spatial heterogeneity of image content. This content-agnostic quantization creates a mismatch with the diffusion model's noise-dependent prior: textured regions receive insufficient generative intervention while smooth regions are over-regularized, limiting the overall rate-perception performance. Second, these codecs suffer from an \textit{information concentration bottleneck} caused by the architectural mismatch between the high-dimensional noisy latent and the fixed 4-channel input of the pre-trained diffusion decoder~\cite{rombach2022high}. Without explicit supervision, the model may fail to concentrate essential semantic information into the primary channels, leading to a non-adaptive latent representation that does not prioritize critical content. Third, existing methods struggle with \textit{ineffective textual conditioning}: they either incur substantial bitrate overhead by transmitting textual side information or rely on generic prompts that lack content relevance, failing to provide content-adaptive textual guidance without bitrate costs.\par

To tackle the aforementioned limitations and establish a content-adaptive diffusion-based image codec, we introduce three technical innovations that address the identified limitations. First, we propose an Uncertainty-Guided Adaptive Quantization (UGAQ) method that fundamentally redesigns the quantization process. Our key innovation lies in learning a spatially-varying uncertainty map from the residual between the main latent representation and the upsampled hyperprior latent. This uncertainty map modulates the quantization noise level across different spatial locations, ensuring that textured regions receive stronger generative intervention while smooth regions maintain structural fidelity. This method effectively aligns the quantization-induced distortion with the diffusion model's noise-dependent denoising strategy, enabling content-aware noise shaping. Second, we develop an Auxiliary Decoder-Guided Information Concentration (ADGIC) method that explicitly solves the information concentration bottleneck. We introduce a lightweight auxiliary decoder that operates exclusively on the first four channels of the noisy latent, producing an auxiliary reconstruction and computing its distortion against the original image. This design forces the compression model to concentrate semantically critical information into the primary channels used by the diffusion decoder, ensuring that the essential visual content is properly preserved and accessible to the generative prior, thereby enforcing a content-driven information allocation. Third, we design a Bitrate-Free Adaptive Textual Conditioning (BFATC) method that enables effective textual conditioning without textual bitrate overhead. We generate content-adaptive captions using a pre-trained BLIP model~\cite{li2022blip} from the auxiliary reconstruction image, providing semantically meaningful guidance to the diffusion process while requiring zero additional textual bitrate, effectively bridging the gap between conditioning quality and bitrate efficiency and achieving dynamic, content-specific conditioning.\par
In summary, our contributions are as follows:

\begin{itemize}
\item We identify three key limitations in current diffusion-based image codecs that hinder content adaptation:~i) the mismatch from content-agnostic isotropic quantization,~ii) the non-adaptive latent representation due to the information concentration bottleneck, and iii) the inability to provide content-aware textual guidance efficiently.

\item We propose three novel methods that collectively establish a content-adaptive diffusion-based image codec: Uncertainty-Guided Adaptive Quantization for content-aware noise shaping, Auxiliary Decoder-Guided Information Concentration for content-driven information allocation, and Bitrate-Free Adaptive Textual Conditioning for content-specific semantic guidance.

\item We validate our codec across various datasets and evaluation criteria, showing improved quantitative and qualitative results, particularly at ultra-low bitrates.
\end{itemize}

\section{Related Work}
\label{sec:related_work}

\subsection{Learned Image Compression}
Learned image compression has gained significant research interest in recent years. Mosts methods typically build upon three core components: non-linear transforms, quantization, and entropy models.
Early work by Ballé~\etal~\cite{DBLP:conf/iclr/BalleLS17} established a foundational framework using convolutional networks with generalized divisive normalization (GDN) for non-linear analysis and synthesis transforms. To enable gradient-based training, they approximated quantization with additive uniform noise during optimization, combined with a factorized entropy model. Subsequent efforts have focused on enhancing each of these components. For transform design, attention mechanisms~\cite{cheng2020image,feng2025linear} and transformer architectures~\cite{zou2022devil,lu2021transformer,liu2023learned} have been incorporated to better capture both global structures and local textures.
For quantization, several alternatives have been proposed to improve quantizer differentiability and efficiency while maintaining inference-time discretization, including soft-to-hard quantization~\cite{agustsson2017soft}, soft-then-hard quantization~\cite{guo2021soft}, universal quantization~\cite{agustsson2020universally}, and non-uniform quantization~\cite{ge2024nlic}. 
Entropy modeling has also seen considerable advances. Early hyperprior models~\cite{DBLP:conf/iclr/BalleMSHJ18} introduced side information to capture spatial dependencies, while autoregressive model~\cite{DBLP:conf/nips/MinnenBT18} further improved accuracy through spatial conditioning. More recent methods combine these ideas, such as channel-wise autoregressive models~\cite{minnen2020channel}, checkerboard context models~\cite{he2021checkerboard}, and hybrid designs~\cite{he2022elic,fu2024fast,jiang2023mlic}. Transformer-based entropy models~\cite{qian2022entroformer,li2024groupedmixer} have also been explored to leverage long-range dependencies for more accurate probability estimation.
\subsection{Diffusion-Based Image Compression}
The success of diffusion models in image generation has motivated their recent adoption in extreme image compression, enabling highly realistic reconstruction under ultra-low bitrates. These methods typically leverage generative priors from pre-trained diffusion models by conditioning the decoding process on the noisy latent representation extracted from the input image. For example, PerCo~\cite{careil2023towards} fine-tunes a diffusion model using vector-quantized spatial features and global text descriptions generated by BLIP-2~\cite{li2023blip}. ResULIC~\cite{ke2025ultra} further analyzes semantic residuals between the original and reconstructed images, transmitting text-guided residuals to steer the diffusion process. DiffEIC~\cite{li2024towards} and its extension RDEIC~\cite{li2024diffusion} demonstrate that conditioning solely on VAE-compressed latents can yield competitive performance without textual input. Alternatively, Relic~\etal~\cite{relic2024lossy} and later work~\cite{relic2025bridging} model quantization errors in the latent space as noise and recover the image via a diffusion denoising process. Despite gains in perceptual realism, such methods often incur high decoding latency due to multi-step sampling strategies like DDIM~\cite{song2020denoising}. To mitigate this, recent methods~\cite{zhang2025stablecodec,xue2025one, zhang2025ultra, guo2025oscar} have incorporated one-step diffusion models~\cite{sauer2024adversarial}, substantially improving inference efficiency while preserving high perceptual quality. Notwithstanding these advances, several key challenges persist that hinder content adaptation, motivating our work toward further improving compression performance.

\begin{figure*}[t]
  \centering
   \includegraphics[width=0.95\linewidth]{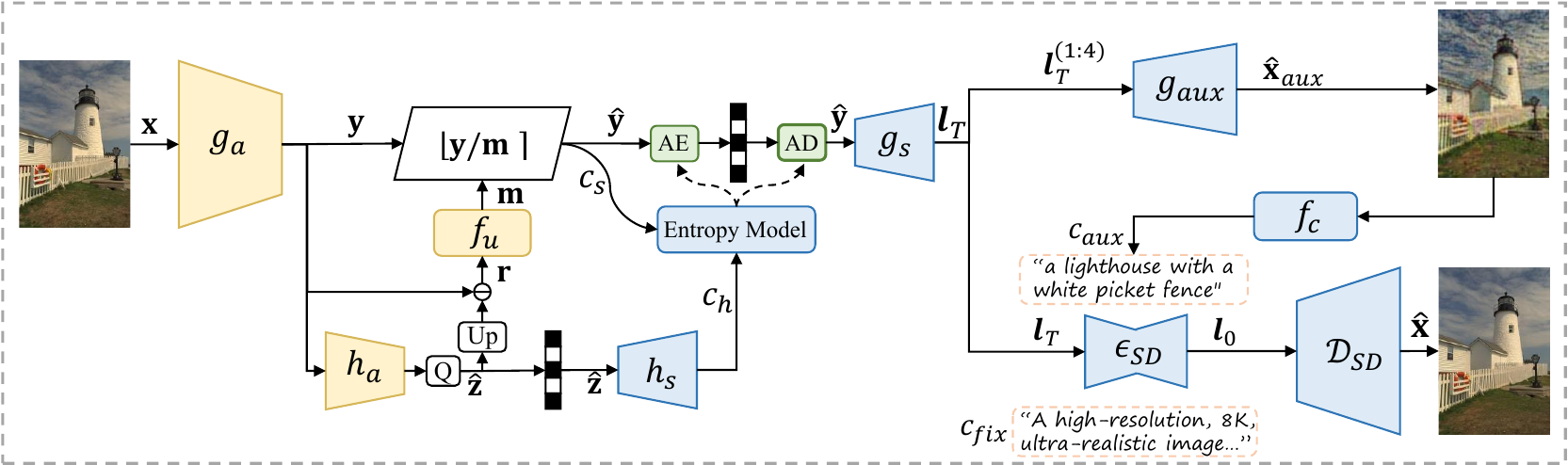}

   \caption{On the encoder side, an analysis transform $g_a$ encodes the input image $\mathbf{x}$ into a compact latent representation $\mathbf{y}$. An uncertainty map $\mathbf{m}$ is estimated by $f_u$ to guide the quantization of $\mathbf{y}$. The quantized latent $\mathbf{\hat{y}}$ is encoded into a bitstream via an arithmetic encoder (AE) and transmitted. On the decoder side, a synthesis transform $g_s$ upsamples $\mathbf{\hat{y}}$ to produce a noisy latent $\bm{l}_T$ at the spatial resolution required by the pre-trained Stable Diffusion VAE decoder $\mathcal{D}_{SD}$~\cite{rombach2022high}. In learned codecs, $\bm{l}_T$ typically has a high channel count (e.g., 320), while $\mathcal{D}_{SD}$ is fixed to accept only 4-channel input. To resolve this, the entire $\bm{l}_T$ is commonly input to the Unet $\epsilon_{S D}$ (a new input channel number is set to the first convolutional layer of the Unet) to utilize all available context for estimating more accurate 4-channel noise (the output channel number of the Unet is still 4)~\cite{zhang2025stablecodec}. The denoising process is applied exclusively to the first four noisy channels $\bm{l}_T^{(1:4)}$, yielding the standard 4-channel clean latent $\bm{l}_0$ for $\mathcal{D}_{SD}$. To concentrate essential semantic information into $\bm{l}_T^{(1:4)}$, a lightweight auxiliary decoder $g_{aux}$ takes $\bm{l}_T^{(1:4)}$ as inputs to reconstruct an auxiliary image $\mathbf{\hat{x}}_{aux}$.  To produce a content-adaptive textual description $c_{aux}$, $\mathbf{\hat{x}}_{aux}$ is captioned by $f_c$ (a frozen BLIP~\cite{li2022blip}). $c_{aux}$ is then combined with a fixed description $c_{fix}$ to condition a one-step diffusion denoising process~\cite{sauer2024adversarial}.}
   \label{fig:framework}
\end{figure*}
\section{Limitations of Diffusion-Based Compression}
\subsection{Isotropic Quantization}~\label{sec:probel_quantization}
A critical yet underexplored challenge in diffusion-based generative compression lies in the mismatch between the content-agnostic quantization strategy and the operational principle of diffusion models. For example, most methods~\cite{li2024towards,guo2025oscar,xue2025one, zhang2025stablecodec} directly quantized the latent representation with a fixed step size, treating the quantization errors as uniform noise to be removed by diffusion process. While Relic~\etal~\cite{relic2025bridging} introduced universal quantization to better align the quantization errors with Gaussian diffusion noise schedule through a derived signal-to-noise ratio (SNR)-matching scheme, it still relies on a global isotropic quantization parameter that is applied uniformly across the entire compact latent representation. This common practice of isotropic quantization implicitly assumes that the distortion introduced by quantization---which the diffusion model is tasked with denoising---affects all image regions equally. However, this assumption contradicts both the nature of image content and the behavior of diffusion models. Natural images exhibit significant spatial heterogeneity, comprising textured regions rich in high-frequency details and smooth regions with minimal information. Diffusion models, in turn, possess noise-level-dependent priors: at high noise levels, they excel at hallucinating realistic textures from a strong generative prior, while at lower noise levels, they act more as conservative denoisers, preserving transmitted structural information. Global isotropic quantization forces a single, compromise noise level upon the entire compact latent representation, failing to adapt to the varying content complexity. Consequently, textured regions suffer from insufficient generative intervention, leading to blurred details, while smooth regions are subjected to unnecessary and potentially disruptive artificiality. This isotropic quantization strategy fails to harness the full potential of the diffusion model's adaptive denoising capabilities due to its inability to perform content-aware noise shaping, creating a suboptimal alignment between the encoded signal and the decoder's generative prior.

\subsection{Information Concentration Bottleneck}~\label{sec:probel_bottleneck}
Diffusion-based image codecs face a fundamental information concentration bottleneck that limits its efficiency and prevents content-aware representation learning. This bottleneck arises from the architectural mismatch between the high-dimensional noisy latents produced by learned codecs and the fixed-dimensional input expected by the pre-trained Stable Diffusion VAE decoder~\cite{rombach2022high}. As shown in~\cref{fig:framework}, learned codecs commonly generate a noisy latent $\bm{l}_T$ with channel dimensions significantly exceeding the standard 4-channel latent space of the VAE decoder. However, only the first four channels $\bm{l}_T^{(1:4)}$ are denoised and fed into the subsequent VAE decoder $\mathcal{D}_{SD}$. This creates a critical constraint: the most semantically meaningful information must be concentrated within $\bm{l}_T^{(1:4)}$ for effective reconstruction. Without explicit guidance, the learning process may fail to adaptively concentrate essential semantic information into the critical first four channels based on image content. The diffusion model's generative prior cannot be fully leveraged if the noisy latent lacks the necessary information density in its primary channels, resulting in a non-adaptive latent representation that fails to preserve content-specific details.

\subsection{Ineffective Textual Conditioning}
Despite the remarkable capability of diffusion models to leverage textual guidance for high-quality image generation, existing diffusion-based compression methods struggle to effectively integrate content-adaptive text conditioning in a rate-efficient manner. Several methods~\cite{careil2023towards,gao2025exploring,ke2025ultra} leverage multimodal large models to automatically generate textual descriptions, which are then losslessly compressed and transmitted as side information. While this provides content-aware guidance for the diffusion denoising process, the associated textual bitrate constitutes some overheads in ultra-low bitrate scenarios.~For example, for the mobile satellite image communication application, the transmission packet size is limited (e.g., 450 bytes). Even low-bitrate texts consume a portion of the precious bit budget, leaving fewer bits for image transmission. Therefore, Zhang~\etal~\cite{zhang2025stablecodec} proposed to avoid transmitting textual descriptions and instead condition the diffusion model on a fixed, generic prompt such as ``\emph{A high-resolution, 8K, ultra-realistic image with sharp focus, vibrant colors, and natural lighting}". Although this eliminates textual bitrate cost, the prompt is inherently agnostic to image content, failing to provide semantically meaningful guidance and limiting the model's ability to reconstruct content-specific details. The fundamental limitation of these methods is their inability to provide content-adaptive textual conditioning without incurring additional textual bitrate costs, thereby hindering the potential for semantically-aware reconstruction.

\section{Method}
\subsection{Uncertainty-Guided Adaptive Quantization}
To address the limitations of content-agnostic global isotropic quantization in diffusion-based compression, we propose an Uncertainty-Guided Adaptive Quantization method that aligns the quantization process with the diffusion model's noise-dependent generative prior, thereby achieving content-adaptive quantization. \par
Specifically, as illustrated in~\cref{fig:framework}, our method begins by upsampling the hyperprior latent $\mathbf{\hat{z}}$ to match the spatial resolution of the main latent representation $\mathbf{y}$:
\begin{equation}
\mathbf{\bar{z}} = \text{UP}(\mathbf{\hat{z}}),
\end{equation}
where $\text{UP}(\cdot)$ denotes bilinear upsampling. We then compute the residual between $\mathbf{y}$ and $\mathbf{\bar{z}}$:
\begin{equation}
\mathbf{r} = \mathbf{y} - \mathbf{\bar{z}}.
\end{equation}
This residual reflects the uncertainty between $\mathbf{y}$ and $\mathbf{\bar{z}}$---the larger the residual in a region, the less information $\mathbf{\bar{z}}$ conveys about $\mathbf{y}$, indicating more complex image textures and serving as a basis for content-aware adaptation.
The residual is then processed through a lightweight uncertainty estimation network $f_u$ to predict an uncertainty map:
\begin{equation}
\mathbf{m} = f_u(\mathbf{r}),
\end{equation}
where each element of $\mathbf{m}$ is restricted to be  $m_{i,j} \geq 1$. \par
We then modulate the latent representation $\mathbf{y}$ using the uncertainty map $\mathbf{m}$ to achieve content-adaptive scaling:
\begin{equation}
\mathbf{\bar{y}} = \mathbf{y} / \mathbf{m},
\end{equation}
where $``/"$ denotes element-wise division. The modulated latent $\mathbf{\bar{y}}$ is then quantized:
\begin{equation}
\mathbf{\hat{y}} = Q(\mathbf{\bar{y}}) = \left\lfloor \frac{\mathbf{\bar{y}}}{\Delta} \right\rceil \cdot \Delta,
\end{equation}
where $Q(\cdot)$ represents the quantization function, $\lfloor \cdot \rceil$ denotes rounding to the nearest integer, and $\Delta$ is the quantization bin width. The quantization process can be approximated as the addition of independent uniform noise:
\begin{equation}
\mathbf{\hat{y}} - \mathbf{\bar{y}} \approx \boldsymbol{\epsilon}, \quad \epsilon_{i,j} \sim \mathcal{U}(-\Delta/2, \Delta/2),
\end{equation}
where $\boldsymbol{\epsilon}$ represents the quantization error approximated as i.i.d. uniform noise, and $\mathcal{U}(-\Delta/2, \Delta/2)$ denotes a uniform distribution over the interval $[-\Delta/2, \Delta/2]$. The quantized latent representation is therefore:
\begin{equation}
\mathbf{\hat{y}} \approx \mathbf{\bar{y}} + \boldsymbol{\epsilon} = \mathbf{y} / \mathbf{m} + \boldsymbol{\epsilon}.
\end{equation}

At the decoder, $\mathbf{\hat{y}}$ is directly feed into the diffusion model without applying the inverse scaling.  
This formulation reveals that although the quantization noise $\boldsymbol{\epsilon}$ has a constant variance $\sigma_{\epsilon}^2 = \Delta^2/12$~\cite{relic2025bridging}, the pre-quantization modulation by $\mathbf{m}$ creates a locally varying signal-to-noise ratio (SNR) at the decoder input, enabling content-aware noise shaping. The effective local SNR for a latent element $y_{i,j}$ can be characterized as:
\begin{equation}
\text{SNR}_{i,j} \propto \frac{\mathbb{E}[\bar{y}_{i,j}^2]}{\sigma_{\epsilon}^2} = \frac{\mathbb{E}[y_{i,j}^2]}{m_{i,j}^2 \cdot \sigma_{\epsilon}^2},
\end{equation}
where $\mathbb{E}[\bar{y}_{i,j}^2]$ is the power of the modulated latent.~This quantitative relationship leads to our central content-adaptive mechanism:

\begin{itemize}
\item High uncertainty regions ($m_{i,j}$ is large):~The signal power is reduced by a factor of $m_{i,j}^2$, resulting in low local SNR. The diffusion model therefore relies more heavily on its generative prior to synthesize details, adaptively enhancing texture generation where needed.

\item Low uncertainty regions ($m_{i,j}$ is small): The signal power remains largely unchanged, maintaining high local SNR. The diffusion model thus prioritizes faithful preservation of the transmitted structural information, adaptively conserving fidelity in smooth areas.
\end{itemize}

Our UGAQ method thereby reduces the misalignment between the quantization distortion and the diffusion model's noise-dependent denoising strategy through content-aware noise shaping, adaptively leveraging its generative capabilities based on local image content.

\subsection{Auxiliary Decoder-Guided Information Concentration}\label{sec:auxiliary_decoder}
To address the information concentration bottleneck and enable content-aware latent representation learning, we propose an Auxiliary Decoder-Guided Information Concentration method.
Our key insight is that without explicit supervision, it may be difficult for the first four channels of the noisy latent contain essential semantic information for high-quality reconstruction, resulting in a non-adaptive latent distribution. Therefore, we introduce a lightweight auxiliary decoder $g_{aux}$ that operates exclusively on these primary channels, providing direct supervision to optimize their information-carrying capacity and enforce content-driven information allocation.\par

Formally, given the full noisy latent $\bm{l}_T$ with $C$ channels where $C \gg 4$, as shown in~\cref{fig:framework}, we feed its first four channels $\bm{l}_T^{(1:4)}$ into an auxiliary decoder $g_{aux}$ to produce an auxiliary reconstructed image $\mathbf{\hat{x}}_{aux}$:
\begin{equation}
\mathbf{\hat{x}}_{aux} = g_{aux}(\bm{l}_T^{(1:4)}).
\end{equation}

We then compute an auxiliary reconstruction loss between this output and the original image $\mathbf{x}$:
\begin{equation}
\mathcal{L}_{aux} = \|\mathbf{x} - \mathbf{\hat{x}}_{aux}\|_2^2.
\end{equation}
This auxiliary reconstruction loss will be incorporated into the overall loss function~\cite{zhang2025stablecodec} to optimize the model.

\begin{figure*}[t]
  \centering
   \includegraphics[width=0.9\linewidth]{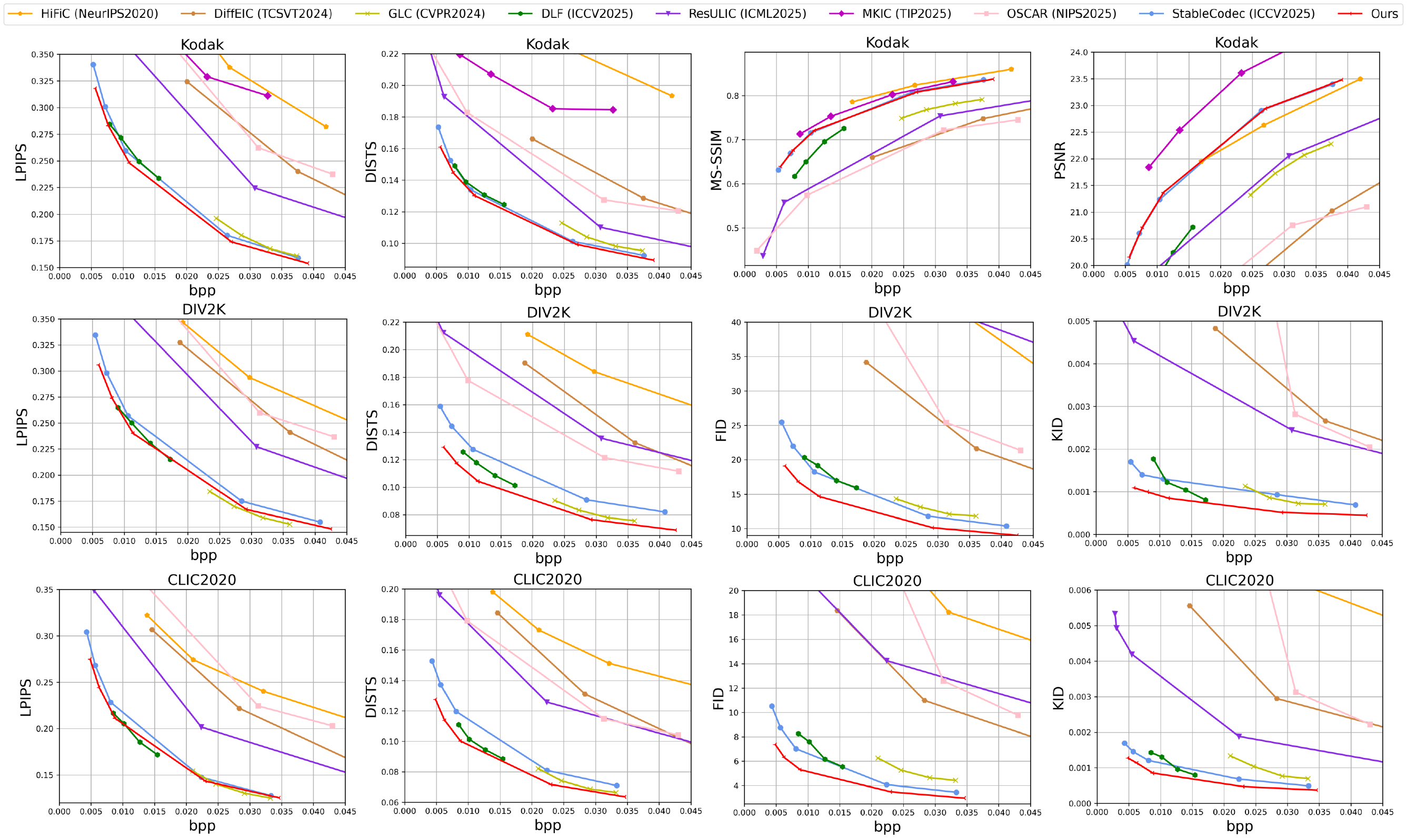}

   \caption{Quantitative comparisons of different generative image codecs on  Kodak, DIV2K Val, and CLIC 2020 Test.}
   \label{fig:RD}
\end{figure*}
\subsection{Bitrate-Free Adaptive Textual Conditioning}\label{sec:method_textual_conditioning}
To overcome the limitation of being unable to obtain content-adaptive textual conditioning without additional textual bitrate costs, we propose a Bitrate-Free Adaptive Textual Conditioning method. Our key insight is to leverage the auxiliary reconstruction $\mathbf{\hat{x}}_{aux}$---generated by the lightweight auxiliary decoder $g_{aux}$ introduced in Section~\ref{sec:auxiliary_decoder}---as a proxy to infer a content-aware textual description. Since $\mathbf{\hat{x}}_{aux}$ is derived entirely from the noisy latent $\bm{l}_T^{(1:4)}$, our method  provides semantically meaningful textual guidance with zero textual bitrate cost.\par

As shown in~\cref{fig:framework}, the auxiliary reconstructed image $\mathbf{\hat{x}}_{aux}$ is fed into an image captioning model $f_c$ (a frozen BLIP~\cite{li2022blip}) to produce a content-adaptive textual description $c_{aux}$:
\begin{equation}
c_{aux} = f_c(\mathbf{\hat{x}}_{aux}).
\end{equation}
To enhance robustness and stability, we then combine $c_{aux}$ with a fixed, generic textual description $c_{fix}$ (``\emph{A high-resolution, 8K, ultra-realistic image with sharp focus, vibrant colors, and natural lighting}")~\cite{zhang2025stablecodec,zhang2024degradation} using the simple string concatenation:
\begin{equation}
c =  c_{aux} + c_{fix}.
\end{equation}
This combined textual description $c$  is used as the conditions for the diffusion denoising process.

\subsection{Implementation}\label{sec:implementation}
Our codec follows the prevalent auto-encoder architecture commonly adopted in learned image compression. To balance generative capability against decoding complexity, we employ a distilled version~\cite{sauer2024adversarial} of Stable Diffusion 2.1~\cite{rombach2022high} to achieve \textbf{one-step diffusion}. A lightweight \emph{blip-image-captioning-base} model is used for auxiliary reconstructed image captioning. For entropy modeling, we integrate both a hyperprior $c_h$ and a spatial prior $c_s$, where the latter is generated by a 4-step autoregressive entropy model with quadtree partitioning~\cite{li2023neural}.
 \par
We train our codec on the training sets of DF2K~\cite{lim2017enhanced} and CLIC 2020 Professional~\cite{toderici2020clic} with the rate-distortion objective:
\begin{equation}
\mathcal{L}=\lambda\mathcal{R}+\mathcal{D},
\end{equation}
where $\lambda$ is the Lagrange multiplier, $\mathcal{R}$ is the bitrate. In addition to our proposed auxiliary reconstruction loss, the distortion term $\mathcal{D}$ incorporates several components from~\cite{zhang2025stablecodec}, including MSE, LPIPS (with VGG features), a CLIP distance~\cite{radford2021learning}, and a GAN loss. More implementation details can be found in the supplementary materials.

\begin{figure*}[t]
  \centering
   \includegraphics[width=0.85\linewidth]{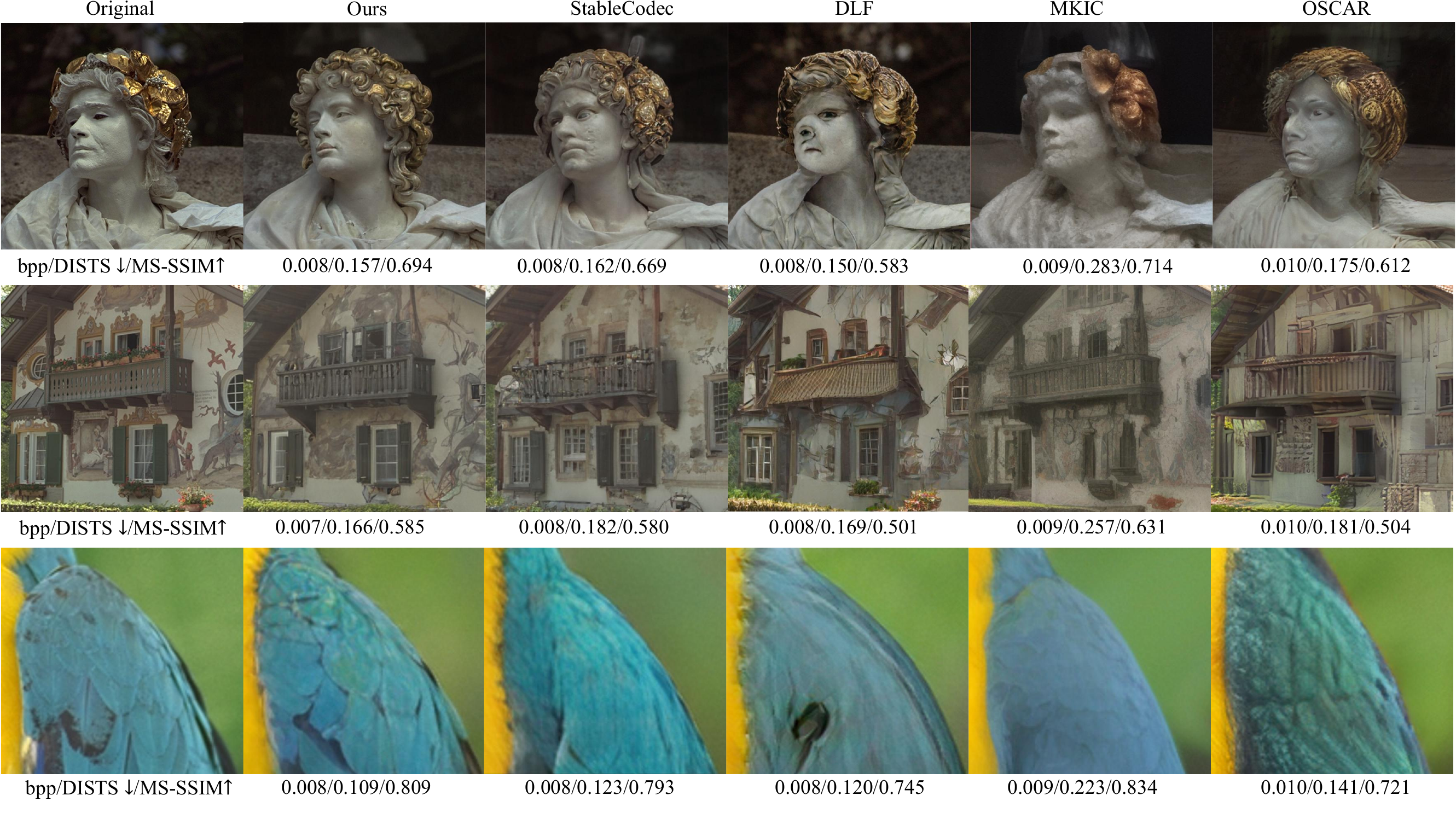}

   \caption{ Qualitative comparison of different generative image codecs on the Kodak dataset under ultra-low bitrate conditions.}
   \label{fig:subjective}
\end{figure*}
\section{Experiments}
\subsection{Experimental Settings}
\textbf{Test Datasets.} We use Kodak~\cite{kodak1993kodak}, the validation set of DIV2K (DIV2K Val)~\cite{agustsson2017ntire}, and the test set of CLIC 2020 Professional (CLIC 2020 Test)~\cite{toderici2020clic} for evaluation. The Kodak dataset contains 24 images with a resolution of $768 \times 512$. The DIV2K Val and the CLIC 2020 Test contain 100 and 428 high-quality 2K-resolution images, respectively. All images are evaluated with the original resolution.\par
\textbf{Evaluation Metrics.}We measure bitrate cost in bits per pixel (bpp). Following prior work~\cite{xue2025dlf,relic2025bridging,zhang2025stablecodec}, we evaluate perceptual quality using DISTS~\cite{ding2020image}, LPIPS~\cite{zhang2018unreasonable} , FID~\cite{heusel2017gans}, and KID~\cite{zhang2024degradation}. Note that FID and KID are omitted on the Kodak dataset due to its limited size.  Results of PSNR and MS-SSIM are also provided in the supplementary material.\par

\textbf{Comparison Methods.} We compare our codec against several generative image codecs, including: HiFiC~\cite{mentzer2020high}, a representative GAN-based codec~\cite{goodfellow2020generative}; DLF~\cite{xue2025dlf} and GLC~\cite{jia2024generative}, which are leading vector quantization-based codecs; and a range of diffusion-based codecs, namely DiffEIC~\cite{li2024towards}, ResULIC~\cite{ke2025ultra}, MKIC~\cite{gao2025exploring}, OSCAR~\cite{guo2025oscar}, and StableCodec (our redesigned variant; see the supplementary material for details)~\cite{zhang2025stablecodec}.

\subsection{Quantitative and Qualitative Comparisons}
As illustrated in~\cref{fig:RD},~we evaluate the quantitative compression performance of different generative image codecs under ultra-low bitrate conditions.~Our codec consistently outperforms other diffusion-based codecs---including DiffEIC, ResULIC, MKIC,  OSCAR, and StableCodec---across all evaluated datasets in terms of LPIPS, DISTS, FID, and KID, validating the effectiveness of our design.
We further provide a qualitative comparison on the Kodak dataset in \cref{fig:subjective} under comparable bitrates. Visual results show that competing codecs such as StableCodec, DLF, MKIC, and OSCAR produce overly smooth regions or lose fine-grained textures. In contrast, our reconstructions preserve more high-frequency details and exhibit more natural visual characteristics. More results can be found in the supplementary material.

\begin{table}[t]
\caption{Ablation studies of our proposed methods on Kodak. Negative BD-rate (\%) values indicate better compression performance. The distortion is measured by LPIPS and DISTS.}
\centering
\scalebox{0.95}{
\begin{tabular}{c|c|c|c|c|c}
\toprule[1.5pt]
Models                & UGAQ & ADGIC & BFATC & LPIPS & DISTS \\ \hline
$M_0$ & \XSolidBrush & \XSolidBrush & \XSolidBrush & 0.00   & 0.00       \\ \hline
$M_1$                    & \Checkmark & \XSolidBrush & \XSolidBrush & --3.7  & --2.7      \\ \hline
$M_2$                    & \Checkmark & \Checkmark & \XSolidBrush & --5.3  & --3.5      \\ \hline
$M_3$                    & \Checkmark & \Checkmark & \Checkmark & --6.8  & --5.5     \\
\bottomrule[1.5pt]
\end{tabular}
}
\label{table:ablation}
\end{table}
\subsection{Ablation Study}
In this section, we conduct ablation studies to validate the effectiveness of our proposed methods.\par
\begin{figure}[t]
  \centering
   \includegraphics[width=0.95\linewidth]{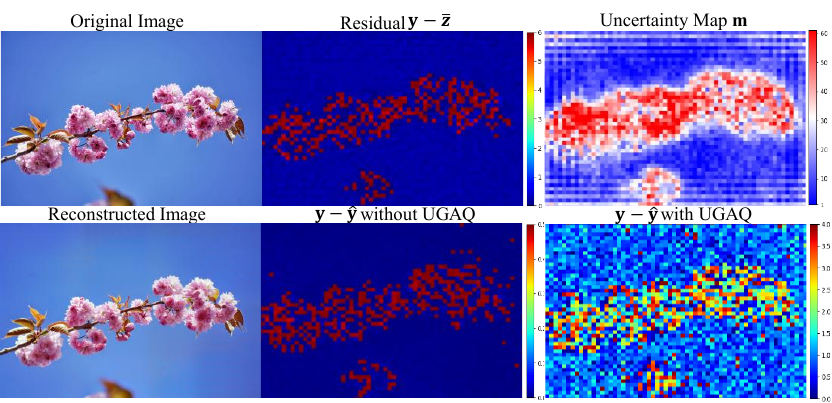}

   \caption{Analysis of Uncertainty-Guided Adaptive Quantization (UGAQ) and the isotropic quantization on the DIV2K dataset.}
   \label{fig:ablation_Uncertainty_Quantization}
\end{figure}

~\textbf{Uncertainty-Guided Adaptive Quantization.}~We first evaluate the contribution of our UGAQ method by integrating it into the baseline Model $M_0$ to construct Model $M_1$. As shown in~\cref{table:ablation}, UGAQ yields BD-rate reductions of 3.7\% and 2.7\% on LPIPS and DISTS, respectively.~This improvement is attributed to the ability of UGAQ to provide content-aware quantization, in contrast to isotropic quantization which applies uniform quantization strength regardless of local content complexity. As visualized in~\cref{fig:ablation_Uncertainty_Quantization}, the residual $\mathbf{y}-\mathbf{\bar{z}}$ serves as an indicator of content uncertainty, with larger values corresponding to highly textured regions. Our method translates this signal into a spatially-varying uncertainty map $\mathbf{m}$, where regions with more complex texture are assigned larger values---opposing existing spatial scaling-based quantization method~\cite{li2022hybrid}. This map guides the quantization process in a content-adaptive manner, producing quantization residuals $\mathbf{y}-\mathbf{\hat{y}}$ that exhibit clear semantic alignment and strongly correlate with content complexity. This demonstrates the ability of UGAQ to actively shape quantization errors, thereby aligning the quantization distortion with the diffusion model's noise-dependent denoising behavior through content-aware noise allocation.

\begin{figure}[t]
  \centering
   \includegraphics[width=0.7\linewidth]{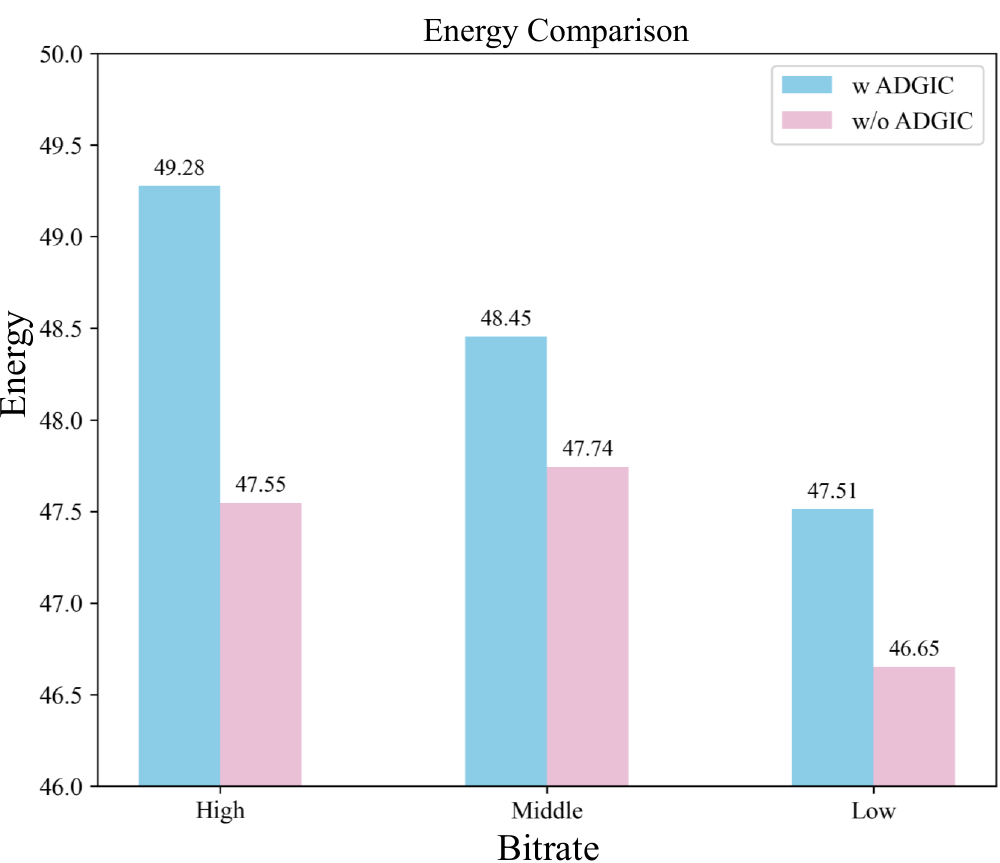}

   \caption{Energy comparison of the first four channels of the noisy latents of the codecs with and without Auxiliary Decoder-Guided Information Concentration (ADGIC) on the Kodak dataset.}
   \label{fig:ablation_energy_comparison}
\end{figure}

\textbf{Auxiliary Decoder-Guided Information Concentration.}~We proceed to validate our ADGIC method by progressively integrating it into Model $M_1$.~The results in~\cref{table:ablation} show that ADGIC yields a further BD-rate reduction of 1.6\% (from $-3.7\%$ to $-5.3\%$) on LPIPS and 0.8\% (from $-2.7\%$ to $-3.5\%$) on DISTS.
To better understand its working mechanism, we analyze the energy distribution across the first four channels of the noisy latent $\bm{l}_T^{(1:4)}$ with and without ADGIC under varying bitrate conditions on the Kodak dataset. Following~\cite{cheng2019learning}, we employ variance as a measure of channel energy. As shown in \cref{fig:ablation_energy_comparison}, ADGIC enhances energy concentration in the primary channels, demonstrating its role in enforcing content-aware information allocation by prioritizing essential semantics adaptively.

\begin{figure}[t]
  \centering
   \includegraphics[width=0.67\linewidth]{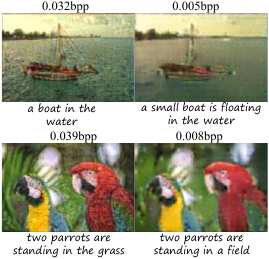}

   \caption{Illustration of the textual descriptions extracted from auxiliary reconstructed images under different bitrate conditions.}
   \label{fig:ablation_dynamic_caption}
\end{figure}

\textbf{Bitrate-Free Adaptive Textual Conditioning.}~We further validate our BFATC method by integrating it into Model $M_2$. As reported in~\cref{table:ablation}, BFATC brings an additional 1.5\% BD-rate improvement (from --5.3\% to --6.8\%) on LPIPS and 2.0\% (from --3.5\% to --5.5\%) on DISTS. This performance gain demonstrates that providing content-adaptive textual guidance is crucial for enhancing compression efficiency. 
Our BFATC method remains robust across bitrates. As visualized in~\cref{fig:ablation_dynamic_caption}, even under severe reconstruction noise at low bitrates, the auxiliary reconstructed image retains sufficient semantic content to produce textual descriptions that remain semantically consistent.\par

\section{Conclusion}
In this work, we address three key limitations hindering content adaptation in diffusion-based image compression: the misalignment from isotropic quantization, the information concentration bottleneck in latent representation, and the inefficiency of textual conditioning. To this end, we propose a content-adaptive diffusion-based image codec (CADC) built upon three technical innovations: an Uncertainty-Guided Adaptive Quantization method that aligns effective quantization distortion with content characteristics through content-aware noise shaping; an Auxiliary Decoder-Guided Information Concentration method that ensures essential semantic information is adaptively preserved in the primary latent channels via content-driven allocation; and a Bitrate-Free Adaptive Textual Conditioning method that derives content-specific semantic guidance without textual bitrate overhead. Extensive experiments validate that our codec achieves the state-of-the-art  perceptual quality, particularly at ultra-low bitrates.

\section{Acknowledgement}
This research was supported in part by the HongKong Innovation and Technology Commission (ITC) grant GHP/044/21SZ and PRP/036/24FX, in part by the General Research Fund(GRF) of the RGC of Hong Kong under Grants 11200323, in part by National Natural Science Foundation of China (NSFC)/ResearchGrants Council (RGC) Joint Research Scheme N\_CityU198/24.

{
    \small
    \bibliographystyle{ieeenat_fullname}
    \bibliography{main}
}

 \clearpage
\setcounter{page}{1}
\maketitlesupplementary

\section{Implementation Details}
\label{sec:rationale}
\begin{figure*}[t]
  \centering
   \includegraphics[width=\linewidth]{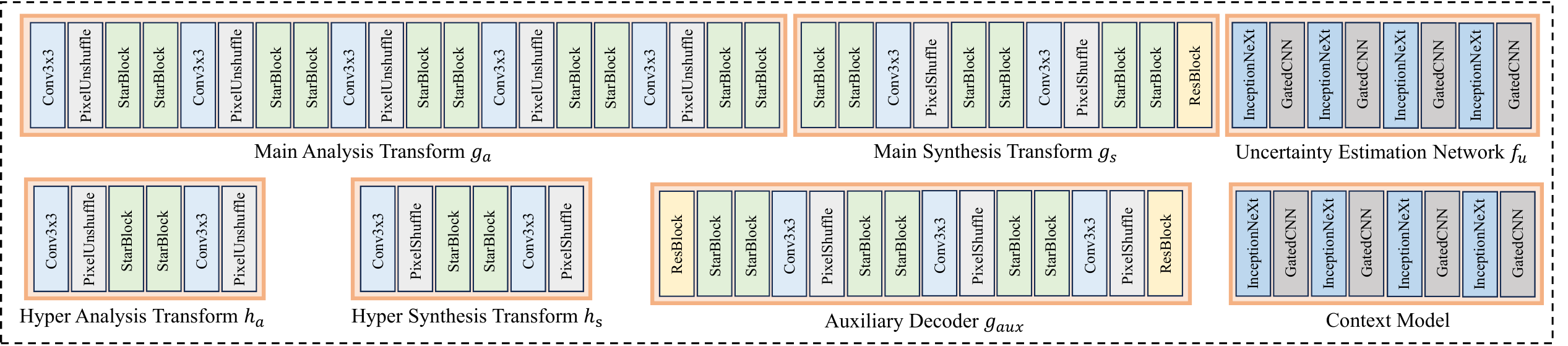}\\
   \includegraphics[width=\linewidth]{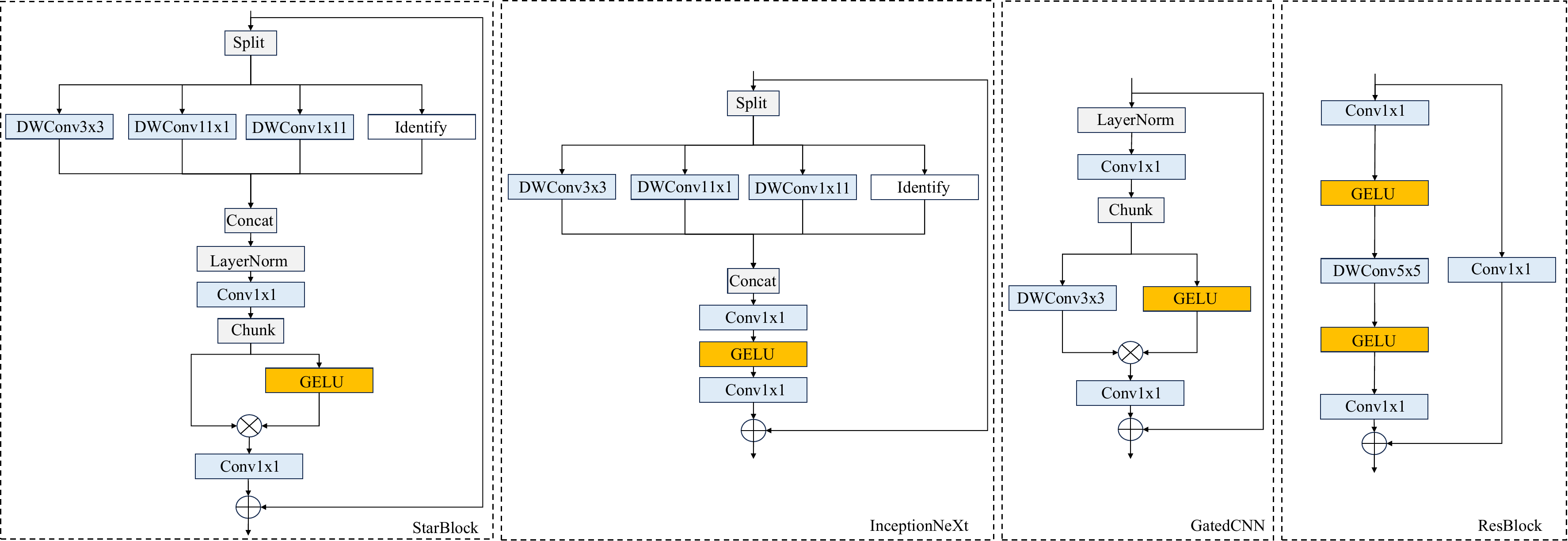}
   \caption{Network structures of the main modules, including the main analysis transform $g_a$, main synthesis transform $g_s$, hyper analysis transform $h_a$, hyper synthesis transform $h_s$, auxiliary decoder $g_{aux}$, uncertainty estimation network $f_u$, and the context model. For efficient network construction, we primarily rely on modified versions of InceptionNeXt~\cite{yu2024inceptionnext}, GatedCNN~\cite{yu2025mambaout}, and their combination (StarBlock).}
   \label{fig:network_structures}
\end{figure*}

\begin{figure}[t]
  \centering
   \includegraphics[width=\linewidth]{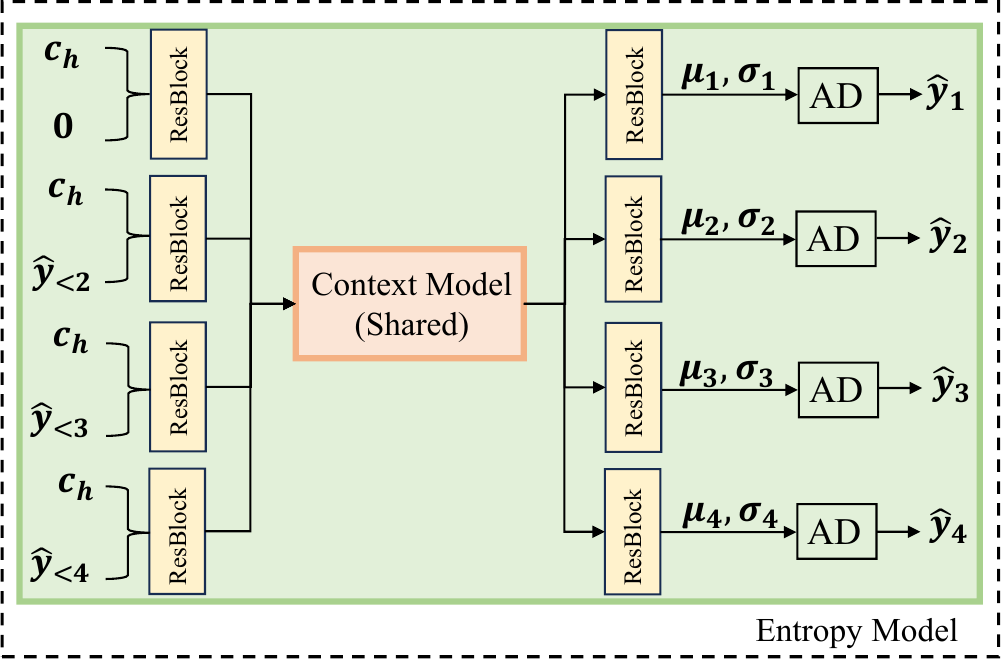}
   \caption{Illustration of our entropy modeling process.}
   \label{fig:entropy_model}
\end{figure}

\begin{figure*}[t]
  \centering
   \includegraphics[width=\linewidth]{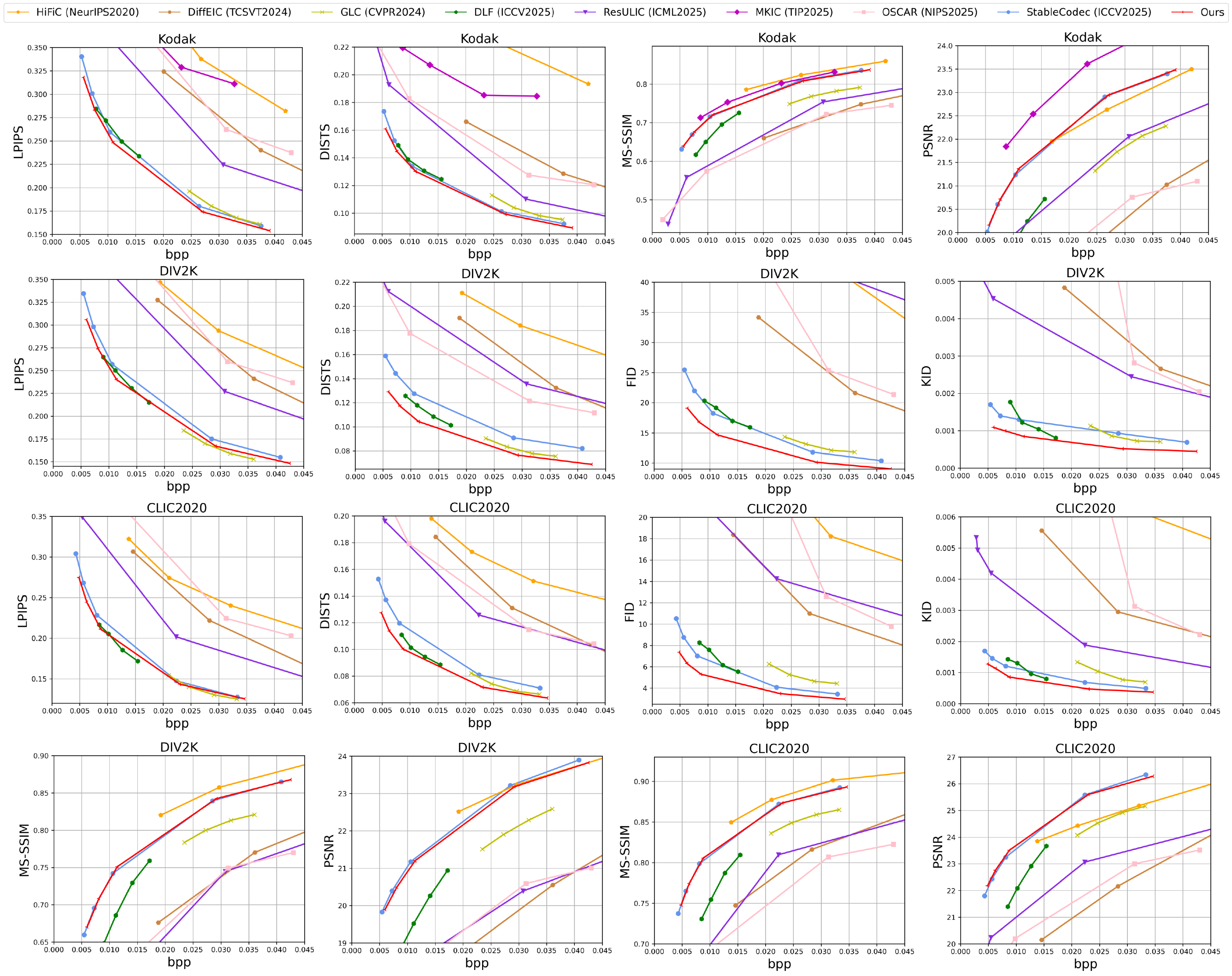}

   \caption{All rate-distortion curves of different generative image codecs on Kodak, the validation set of DIV2K, and the test set of CLIC 2020 Professional in terms of LPIPS, DISTS, FID, KID, MS-SSIM, and PSNR metrics.}
   \label{fig:all_RD}
\end{figure*}

\begin{figure*}[t]
  \centering
   \includegraphics[width=0.68\linewidth]{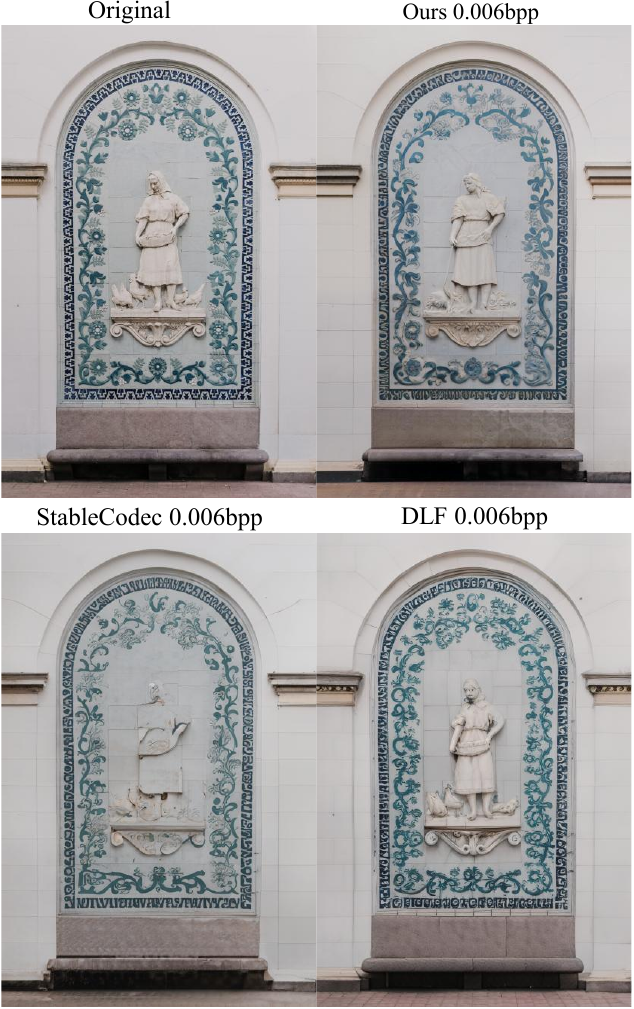}
   \caption{Visual examples and comparisons on 2K-resolution images from the test set of CLIC 2020 Professional.}
   \label{fig:subjective1}
\end{figure*}

\begin{figure*}[t]
  \centering
   \includegraphics[width=0.85\linewidth]{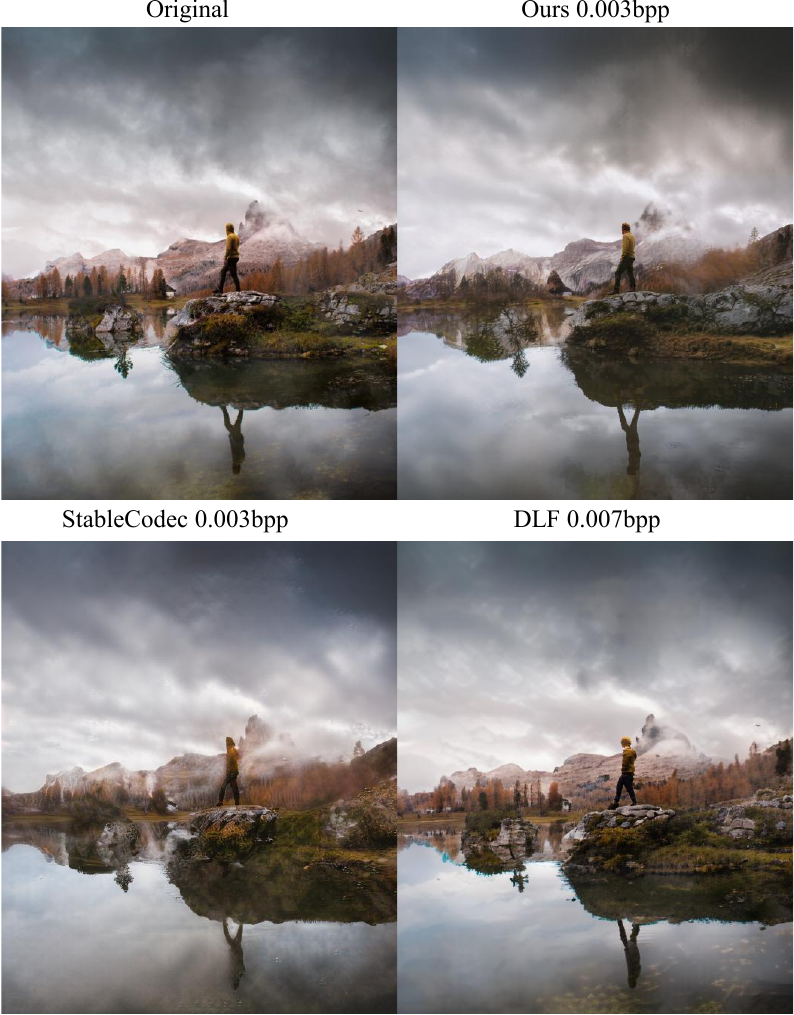}
   \caption{Visual examples and comparisons on 2K-resolution images from the test set of CLIC 2020 Professional.}
   \label{fig:subjective2}
\end{figure*}

\begin{figure*}[t]
  \centering
   \includegraphics[width=\linewidth]{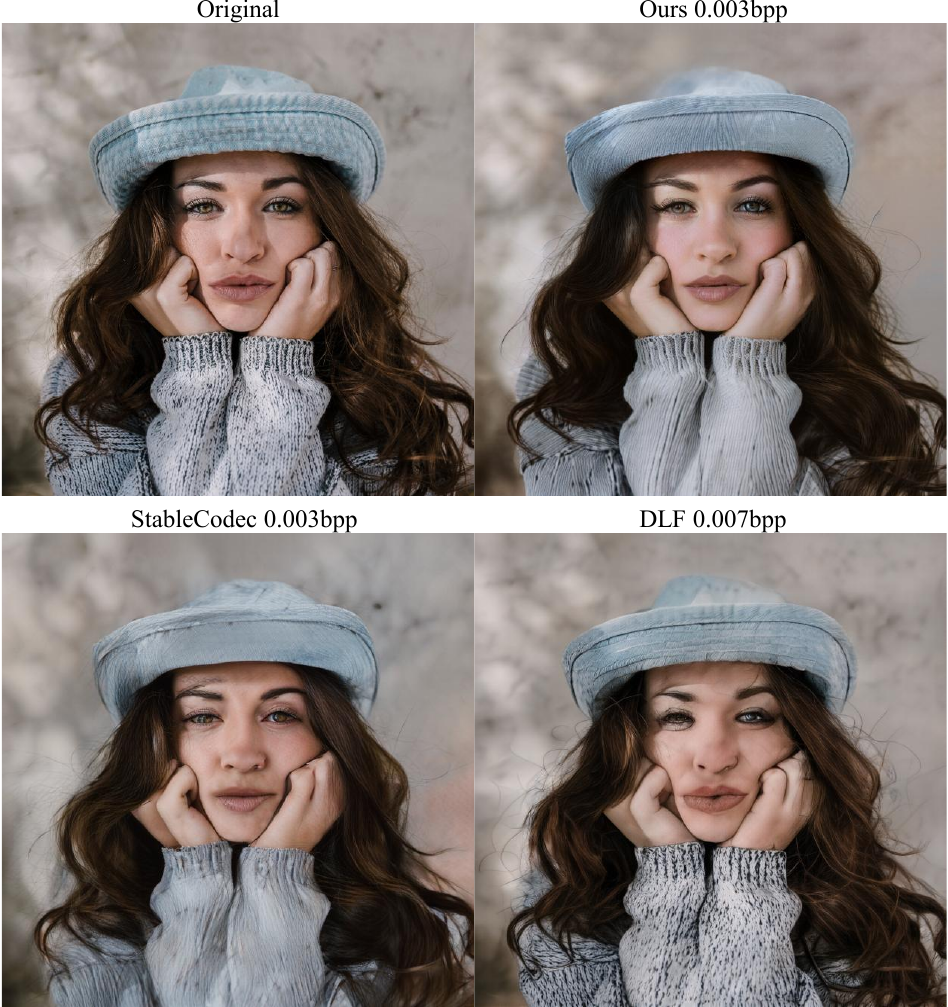}
   \caption{Visual examples and comparisons on 2K-resolution images from the validation set of DIV2K.}
   \label{fig:subjective3}
\end{figure*}

\begin{figure*}[t]
  \centering
   \includegraphics[width=\linewidth]{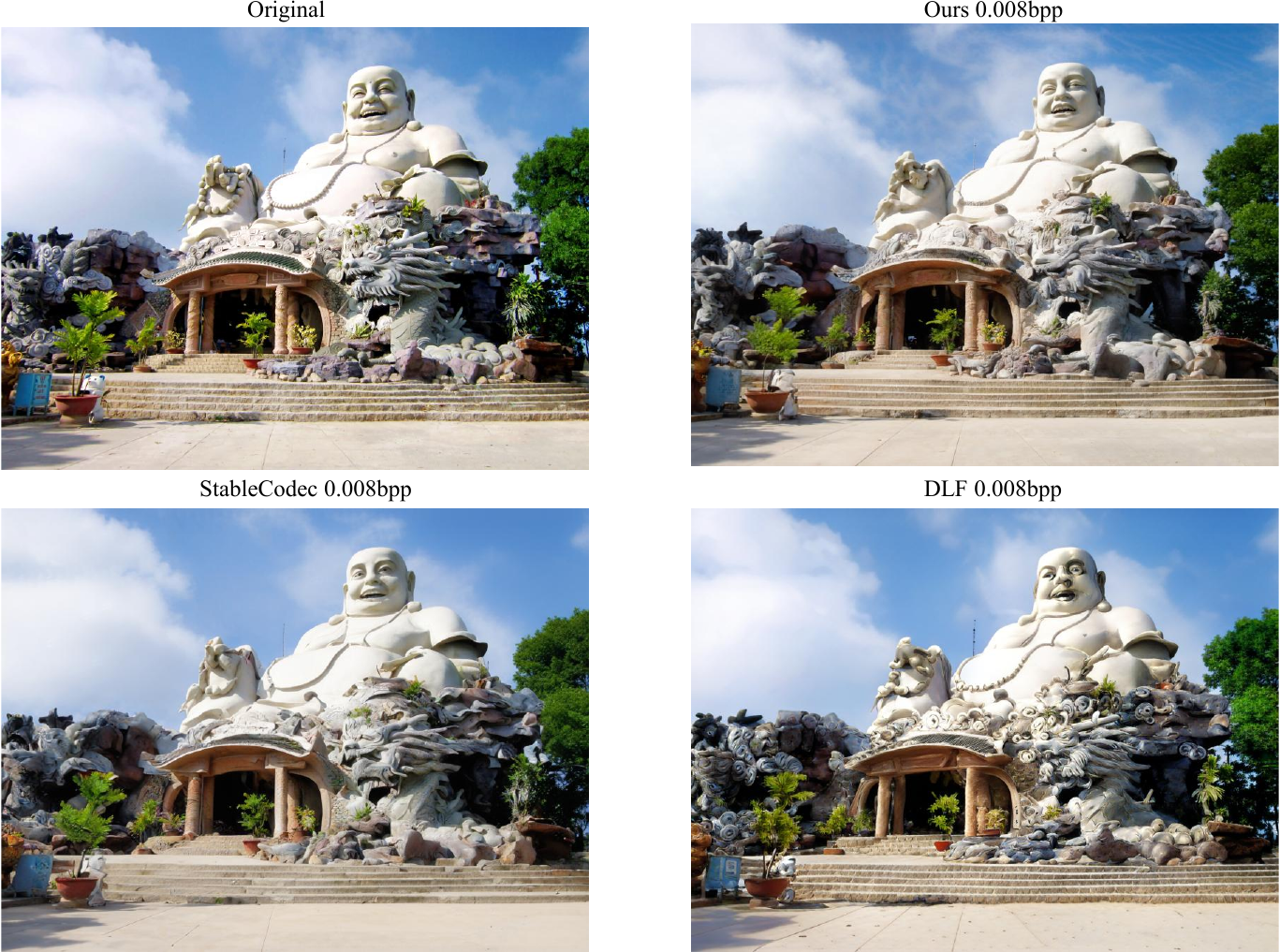}
   \caption{Visual examples and comparisons on 2K-resolution images from the validation set of  DIV2K.}
   \label{fig:subjective4}
\end{figure*}

\subsection{Network Structure}
The detailed architectures of the main modules are illustrated in~\cref{fig:network_structures}, including the main analysis transform $g_a$, main synthesis transform $g_s$, hyper analysis transform $h_a$, hyper synthesis transform $h_s$, auxiliary decoder $g_{aux}$, uncertainty estimation network $f_u$, and the context model. For efficient network construction, we primarily rely on modified versions of InceptionNeXt~\cite{yu2024inceptionnext}, GatedCNN~\cite{yu2025mambaout}, and their combination (StarBlock).\par
We also illustrate the overall entropy modeling process in~\cref{fig:entropy_model}.
Our entropy model integrates a hyperprior module with an autoregressive context model. The hyperprior representation $\mathbf{c_h}$ is derived as:
\begin{equation}
\mathbf{z} = h_a(\mathbf{y}), \quad \mathbf{\hat{z}} = \lfloor \mathbf{z} \rceil, \quad \mathbf{c_h} = h_s(\mathbf{\hat{z}}),
\end{equation}
where $\lfloor \cdot \rceil$ denotes rounding to the nearest integer. Here, $\mathbf{y}$ has 320 channels with a spatial downsampling factor of 64, while $\mathbf{z}$ and $\mathbf{\hat{z}}$ also have 320 channels but with a higher spatial compression ratio of 256$\times$.
To balance coding performance and computational efficiency, we employ a 4-step autoregressive process based on quadtree partitioning~\cite{li2023neural}. As shown in~\cref{fig:entropy_model}, this process estimates the Gaussian parameters $\bm{\mu}$ and $\bm{\sigma}$ for the quantized latent $\mathbf{\hat{y}}$. Subsequently, arithmetic coding is applied to encode or decode $\mathbf{\hat{y}}$ into/from the bitstream.\par
For BLIP, we use the lightweight \emph{blip-image-captioning-base} model (247.41M parameters, with inputs resized to $384\times384$ by default).\par
To enable a fair evaluation of our three proposed methods, we construct a variant StableCodec by removing its original VAE encoder and auxiliary encoder, and replacing the main analysis/synthesis transforms, hyper analysis/synthesis transforms, and context model with our implementations. This reduces computational complexity and eliminates architectural discrepancies that could otherwise confound the assessment of our contributions.  We refer to this variant as our baseline $M_0$.

\subsection{Training Process}
We train our codec based on the standard rate-distortion loss:
\begin{equation}
\mathcal{L}=\lambda\mathcal{R}+\mathcal{D},
\end{equation}
where $\mathcal{R}$ denotes the bitrate, $\mathcal{D}$ is the distortion measure, and $\lambda$ is a Lagrange multiplier that balances the two terms.
We employ a two-stage training strategy~\cite{zhang2025stablecodec}. In the first stage, a base model is trained using a relatively small $\lambda_{\text{base}}$. In the second stage, this pre-trained model is fine-tuned with a larger $\lambda_{\text{target}}$ to achieve ultra-low target bitrates.
The distortion term $\mathcal{D}$ incorporates multiple components: MSE, LPIPS (computed with VGG features), a CLIP-based loss $\mathcal{L}_{\text{CLIP}}$~\cite{zhang2025stablecodec}---defined as the $\ell_2$-distance between CLIP embeddings of $\mathbf{x}$ and $\mathbf{\hat{x}}$, an adversarial loss $\mathcal{L}_{\text{adv}}$, and our proposed auxiliary reconstruction loss $\mathcal{L}_{\text{aux}}$.
We use DINOv2~\cite{oquab2023dinov2} with registers~\cite{darcet2023vision} as the discriminator backbone~\cite{kumari2022ensembling}.  Note that $\mathcal{L}_{\text{adv}}$ and $\mathcal{L}_{\text{aux}}$ are only activated in the second training stage. The complete objective is formulated as follows:
\begin{equation}
\begin{aligned}
& \text { Stage I : } \underset{\theta}{\arg \min } L_1 =  \lambda_{\text {base }} \mathcal{R}+\mathcal{D}_1, \\
& \text { Stage II : } \underset{\theta}{\arg \min } L_2 =  \lambda_{\text {target }} \mathcal{R}+\mathcal{D}_2,\\
& \qquad \begin{aligned}
&\mathcal{D}_1 = d_1 \operatorname{MSE}(x, \hat{x})+d_2 \operatorname{LPIPS}(x, \hat{x})+d_3 \mathcal{L}_{C L I P}(x, \hat{x}) \\
& \begin{aligned}
& \mathcal{D}_2 = d_1 \operatorname{MSE}(x, \hat{x})+d_2 \operatorname{LPIPS}(x, \hat{x})+d_3 \mathcal{L}_{C L I P}(x, \hat{x}) \\
& \quad+d_4 \mathcal{L}_{aux}(x, \hat{x}_{aux})+d_5 \mathcal{L}_{adv},
\end{aligned}
\end{aligned}
\end{aligned}
\end{equation}
where $\theta$ denotes all trainable parameters in the codec, and $d_1$–$d_5$ are weighting coefficients that balance the distortion terms.
\section{More Experimental Results}
\subsection{More Quantitative Results}
A comprehensive evaluation of rate-distortion performance is conducted on Kodak, the validation set of DIV2K~\cite{agustsson2017ntire}, and the test set of CLIC 2020 Professional~\cite{toderici2020clic}, comparing various codecs (HiFiC~\cite{mentzer2020high}, DiffEIC~\cite{li2024towards}, GLC~\cite{jia2024generative}, DLF~\cite{xue2025dlf}, ResULIC~\cite{ke2025ultra}, MKIC~\cite{gao2025exploring}, OSCAR~\cite{guo2025oscar},  StableCodec~\cite{zhang2025stablecodec}) across six metrics: LPIPS, DISTS, FID, KID, MS-SSIM, and PSNR. \Cref{fig:all_RD} shows that our codec consistently delivers state-of-the-art perceptual quality, particularly under ultra-low bitrate conditions.

\subsection{More Qualitative Results}
Following the quantitative analysis, we also provide a more detailed subjective comparison of the three top-performing codecs: our codec, StableCodec~\cite{zhang2025stablecodec}, and DLF~\cite{xue2025dlf}, under ultra-low bitrate conditions. Visual comparisons are presented in \cref{fig:subjective1,fig:subjective2,fig:subjective3,fig:subjective4}. The results show that our codec consistently achieves the best visual quality, corroborating the quantitative findings and confirming its superior perceptual performance at ultra-low bitrates.
\begin{table}[t]
  \centering
\caption{Runtime comparison of in seconds averaged on the Kodak dataset.}
\begin{tabular}{c|c|c}
\toprule[1.5pt]
Method       & Encoding Time (s) & Decoding Time (s) \\ \hline
Ours        & 0.034             & 0.355             \\ \hline
StableCodec & 0.030             & 0.263             \\ \hline
DLF         & 0.181             & 0.195             \\
\bottomrule[1.5pt]
\end{tabular}
\label{table:time}
\end{table}

\subsection{Runtime Comparison}
\Cref{table:time} compares the runtime efficiency of our codec against the variant StableCodec  and DLF on the Kodak dataset using a single RTX 3090 GPU. The encoding time of our codec and the variant StableCodec  is significantly reduced~\cite{zhang2025stablecodec}, as they eliminate the need for the pre-trained VAE encoder used in Stable Diffusion and the auxiliary encoder along with the latent residual prediction (LRP) module. Our proposed uncertainty-guided adaptive quantization only slightly increases the encoding complexity compared to the variant StableCodec. For decoding, since our codec incorporates an auxiliary decoder and a BLIP image captioning model, the overall decoding time brings a moderate increase. In addition, due to the inherent complexity of the diffusion model, our decoding time remains higher than that of DLF. However, it is important to note that in ultra-low bitrate application scenarios, such as mobile-satellite communications, the uplink bandwidth (from the mobile device to the satellite) is often more constrained and valuable than the downlink. This places a premium on low encoding complexity at the client side, while the more computationally intensive decoding can be offloaded to powerful cloud servers. Therefore, our codec---with its fast encoding and competitively performing decoding---retains strong practical value for real-world deployment.

\subsection{User Study}
To comprehensively evaluate the perceptual quality of reconstructed images at ultra-low bitrates, we conduct a user study on the Kodak dataset, comparing our codec with two top-performing codecs: StableCodec~\cite{zhang2025stablecodec} and DLF~\cite{xue2025dlf}. The study follows a top-1 preference protocol, in which participants are asked to select the reconstruction they perceive as most visually consistent with the ground-truth image. Each participant evaluates 24 randomly ordered test cases. For each case, the ground-truth image is displayed together with the three reconstructed images in a single row of four images, with the order of the methods randomized across trials.  Participants are instructed to select the reconstruction that is the most ``consistent" with the original image. A total of 25 participants took part in the study, collectively contributing 600 evaluation cases. The results, summarized in \cref{table:user}, indicate that reconstructions from our codec were preferred in 58.5\% of cases. This strong preference underscores its superior perceptual quality as judged by human observers.
\begin{table}[t]
  \centering
\caption{Top-1 user preference on the Kodak dataset.}
\begin{tabular}{c|c|c|c}
\toprule[1.5pt]
Method                  & Ours & StableCodec & DLF \\ \hline
Bitrate (bpp)           & 0.0076 & 0.0072      & 0.0079     \\ \hline
Top-1  Percentage (\%) &  58.5    &  29.0           & 12.5    \\
\bottomrule[1.5pt]
\end{tabular}
\label{table:user}
\end{table}

\end{document}